\documentclass[letterpaper, 10 pt, conference]{ieeeconf}  

\IEEEoverridecommandlockouts                              

\usepackage{cite}
\usepackage{amsmath,amssymb,amsfonts}
\usepackage{algorithm}
\usepackage[noend]{algpseudocode}
\usepackage{graphicx}
\usepackage{textcomp}
\usepackage{xcolor,colortbl}
\usepackage{booktabs}
\usepackage{multirow}
\usepackage{array}
\usepackage{comment}
\usepackage{subcaption}

\usepackage{svg}
\usepackage[breaklinks,colorlinks,bookmarks,citecolor=flodarkpurple]{hyperref}
\usepackage{siunitx}
\usepackage{caption}
\DeclareCaptionFont{8pt}{\fontsize{8pt}{12pt}\selectfont}
\algrenewcommand\algorithmicindent{0.8em}%

\newcommand{\revised}[1]{#1}

\definecolor{flodarkpurple}{rgb}{0.288,0.1196,0.7}
\definecolor{amber}{rgb}{1.0, 0.75, 0.0}
\definecolor{orange}{rgb}{1.0, 0.58, 0.0}
\definecolor{green}{rgb}{0.0, 0.725, 0.25}
\definecolor{blue}{rgb}{0.047, 0.365, 0.647}
\definecolor{tablegray}{gray}{0.4} 

\newcommand{\mb}[1]{\mathbf{#1}}

\DeclareMathOperator*{\argmax}{arg\,max}

\newcommand{\tb}{T^E}

\newcommand{\pb}{p^{E}}

\newcommand{\etal}{\emph{et al.}}

\newcommand{\authorhref}[3][flodarkpurple]{\href{#2}{\color{#1}{#3}}}
\newcommand{\perpetua}{\textit{Perpetua}}

\def\BibTeX{{\rm B\kern-.05em{\sc i\kern-.025em b}\kern-.08em
    T\kern-.1667em\lower.7ex\hbox{E}\kern-.125emX}}
\begin{document}

\begin{minipage}{\textwidth}

\begin{center}
\vspace*{\fill}
\textcopyright 2025 IEEE.  Personal use of this material is permitted.  Permission from IEEE must be obtained for all other uses, in any current or future media, including reprinting/republishing this material for advertising or promotional purposes, creating new collective works, for resale or redistribution to servers or lists, or reuse of any copyrighted component of this work in other works. 
\vspace*{\fill}
\end{center}

\hfill

This paper has been accepted for publication in the \textit{IEEE/RSJ International Conference on Intelligent Robots and Systems (IROS)}. \\ Please cite this paper as:

\begin{verbatim} 
@article{saavedra2025perpetua,
    	title        = {Perpetua: Multi-Hypothesis Persistence Modeling for Semi-Static 
                     Environments},
    	author       = {Saavedra-Ruiz, Miguel and Nashed, Samer and Gauthier, Charlie and 
                     Paull, Liam},
    	year         = 2025,
    	journal      = {IEEE/RSJ International Conference on Intelligent Robots and 
                     Systems (IROS)},
    	organization = {IEEE}
}  \end{verbatim} 
\end{minipage}
\newpage

\title{Perpetua: Multi-Hypothesis Persistence Modeling for Semi-Static Environments
\\
\thanks{This work was partially funded by an FRQNT B2X Scholarship with DOI 10.69777/335998 [MS], the Natural Sciences and Engineering Research Council of Canada (NSERC), funding reference number 600981-2025 [CG], and under the Discovery Grant Program. Additional support was provided by CIFAR under the CCAI Chairs program.}
\thanks{$^{1}$Department of Computer Science and Operations Research, Université de Montréal, Montréal, QC, Canada.}
\thanks{$^{2}$Mila - Quebec AI Institute, Montréal, QC, Canada.}
}

\author{
\authorhref{https://mikes96.github.io}{Miguel Saavedra-Ruiz}$^{1,2}$, 
\authorhref{https://samernashed.github.io}{Samer B. Nashed}$^{1,2}$, 
\authorhref{https://velythyl.github.io}{Charlie Gauthier}$^{1,2}$, 
\authorhref{https://liampaull.ca}{Liam Paull}$^{1,2}$ \\
\\
}

\makeatletter
\let\@oldmaketitle\@maketitle
\renewcommand{\@maketitle}{\@oldmaketitle
\centering
\includegraphics[width=\linewidth,trim={0.5cm 0 0.5cm 0},clip]{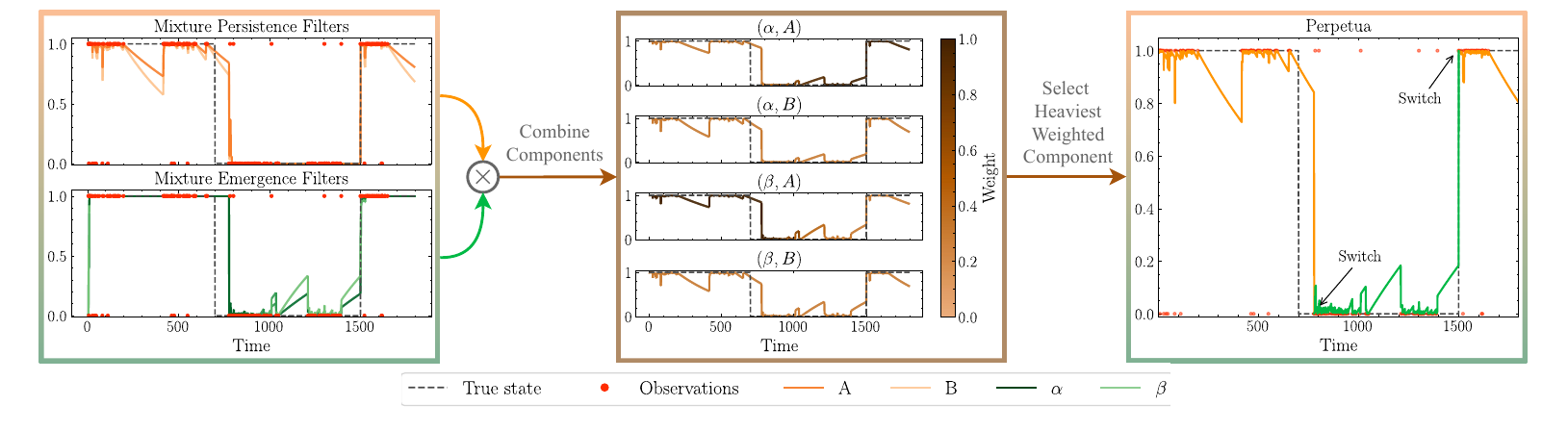}
\vspace{-2.5em}
\setcounter{figure}{0} 
\refstepcounter{figure} 
\addtocounter{figure}{-1} 
\captionof{figure}{
\footnotesize
\textbf{\perpetua} models the presence and absence of semi-static features through the combination of two mixtures models: one for disappearance (mixture of persistence filters) and one for reappearance (mixture of emergence filters), together with a switching mechanism to toggle between them. In the above figure, the mixture of emergence and persistence filters (left panel) each have two components, which combine to make 4 possible outcomes (middle). Based on the available measurements, we select the most likely combination (right) to model the persistence of a semi-static feature.\looseness-1}
\label{fig:splash}
\vspace{-0.98em}
}
\makeatother

\setcounter{figure}{1}  
\addtocounter{figure}{1} 

\maketitle
\thispagestyle{empty}
\pagestyle{empty}

\begin{abstract}

Many robotic systems require extended deployments in complex, dynamic environments. In such deployments, parts of the environment may change between subsequent robot observations. Most robotic mapping or environment modeling algorithms are incapable of representing dynamic features in a way that enables predicting their future state. Instead, they opt to filter certain state observations, either by removing them or some form of weighted averaging. This paper introduces \textit{Perpetua}, a method for modeling the dynamics of semi-static features. Perpetua is able to: incorporate prior knowledge about the dynamics of the feature if it exists, track multiple hypotheses, and adapt over time to enable predicting of future feature states. Specifically, we chain together mixtures of ``persistence'' and ``emergence'' filters to model the probability that features will disappear or reappear in a formal Bayesian framework. The approach is an efficient, scalable, general, and robust method for estimating the states of features in an environment, both in the present as well as at arbitrary future times. Through experiments on simulated and real-world data, we find that Perpetua yields better accuracy than similar approaches while also being online adaptable and robust to missing observations.\looseness-1

\end{abstract}

\section{Introduction}
\label{sec:intro}

Effective robotic planning requires accurate estimation of the robot's operating environment. However, achieving this for both present and future time steps, and for aspects of the environment not currently visible, is particularly challenging due to partial, noisy sensor data and the potential for complex, \emph{semi-static feature dynamics} wherein features (e.g., points, surfels, objects) may appear or disappear between observations. While recent research has shown the value of estimating the state of semi-static features, sometimes called persistence estimation, for tasks such as localization~\cite{adkins2022probabilistic,hashemifar2020visual}, mapping~\cite{schmid2024khronos,nashed2016curating}, navigation~\cite{krajnik2017fremen,qian2024closing,nardi2020nav}, and planning~\cite{krajnik2021khrono}, our ability to estimate persistence in practice remains limited.

Many semi-static changes exhibit a time-dependent nature, such as an office door being open on weekdays and closed on weekends, allowing the possibility of learning persistence estimators that can capture the qualitative dynamics and predict future changes \cite{krajnik2017fremen, wang2020arma}. Ideal world modeling approaches should be: (1) robust to partial observations, (2) adaptable and updatable online, (3) capable of representing multiple dynamic modes (e.g., hourly versus daily changes), and (4) reliant on minimal prior knowledge.

Methods for modeling feature persistence fall on a spectrum between filtering and predictive approaches. Filtering methods, like the persistence filter~\cite{rosen2016towards} and its derivatives~\cite{nobre2018online, hashemifar2020visual}, estimate persistence recursively from observations. These are adaptable and robust to observation noise, but struggle to model feature \emph{reappearance}, and require accurate priors over model parameters to be effective. In contrast, predictive models, such as FreMen~\cite{krajnik2017fremen}, learn a model from data (albeit offline) and use it to predict future persistence. They can model feature reappearance, and capture multiple dynamic modes. However, they cannot adapt to new information online and are susceptible to noise in the training data.\looseness-1

This paper proposes \perpetua~(Fig. \ref{fig:splash}), an efficient method for estimating the persistence of semi-static features that combines the benefits of filtering and predictive approaches, enabling robustness to missing observations, online adaptation from data, future persistence prediction, characterization of multiple dynamical hypotheses, and reduced dependence on prior knowledge. To achieve this, we first develop a posterior inference framework that jointly estimates persistence and latent mixture components in a mixture of persistence filters modeling multiple feature dynamics (\S\ref{sec:death}). Next, we derive an \emph{emergence filter}, based on the persistence filter, to capture feature reappearance (\S\ref{subsec:birth}), and show how to combine mixtures of persistence and emergence filters via a state machine to track and predict the disappearance and reappearance of features arbitrarily far into the future (\S\ref{sec:state_machine}). Finally, we derive an expectation-maximization-based approach for learning model parameters from noisy observations (\S\ref{sec:learning}).\looseness-1

Empirically, Perpetua's persistence estimates are more accurate than baseline methods over a variety of prediction horizons while remaining robust to sensor noise and missing observations. Experiments are conducted on both simulated and real-world datasets exhibiting semi-static changes. \noindent\textbf{Webpage at:} {\small \texttt{\textcolor{flodarkpurple}{https://montrealrobotics.ca/perpetua}}}

\section{Related Work}
\label{sec:rw}

Modeling, tracking, and predicting feature persistence has been an important topic in robotics~\cite{biber2005dynamicMaps,cadena2016past,schmid2022panoptic}. Generally, work in this area falls into one of four broad categories: (1) modeling via Markov chains, (2) using deep learning, (3) predictive models, and (4) filters. Markov chain methods represent presence and absence as states, and model dynamics as state transitions~\cite{saarinen2012markov,tipaldi2013lifelong}. While these approaches can learn parameters from data, they are limited to predicting future persistence at fixed intervals, rather than continuously.

Alternatively, deep learning-based methods have shown an ability to estimate feature persistence, for example, by using object-graphs built from latent representations of point clouds~\cite{fu2023neuse}, or training neural classifiers to predict position, state, and variability of objects on a 3D scene graph~\cite{looper20233d}. Thomas~\etal~\cite{thomas2023foreseeable} employ a kernel point convolution network to predict spatio-temporal occupancy over short time windows. However, except for Thomas~\etal, these methods cannot predict future persistence and many require large amounts of labeled data for training.

Many approaches use either predictive models or filters. Predictive models often relax the assumption of known priors over feature dynamics, and one of the most well-known predictive models is FreMen, introduced by Krajník \etal~\cite{krajnik2017fremen}, which models feature persistence using Fourier analysis enabling future persistence estimates via an inverse Fourier transform. Later, FreMen was extended to jointly model persistence over space and time~\cite{krajnik2019warped}. Other methods have proposed using Guassian processes over dynamic Hilbert maps for continuous future occupancy prediction~\cite{guizilini2019hilbert}, or employed autoregressive moving average models (ARMA) to estimate and predict persistence~\cite{wang2020arma, wang2024arma}. Although these methods offer strong predictive capabilities, they lack online adaptability and rely on observation sequences for training.

Filtering methods, particularly the persistence filter~\cite{rosen2016towards} and its derivatives (e.g.~\cite{nobre2018online,deng2023global}), have several important strengths, including robustness to noisy observations, real-time adaptation, and a strong theoretical foundation. While they can predict future persistence, these models generally become inaccurate quickly. Deng \etal~\cite{deng2023global} addressed this by introducing a long short-term exponential model that updates persistence estimates by prioritizing the latest observation, but this approach remains sensitive to observation noise. The main drawbacks of such methods are the reliance on accurate \textit{a priori} model parameters, a single persistence hypothesis per feature, and the inability to handle feature reappearance.

Perpetua unifies key aspects of these methods: the adaptability and noise resilience of online methods with the predictive power and robustness to priors of offline methods. We extend the persistence filter to a mixture model to track multiple dynamic modes, introduce the emergence filter to model reappearance, present a method for switching between filters, and derive an algorithm to learn model parameters from data.\looseness-1

\section{Background and Preliminaries}

\subsection{Problem Definition}

We consider an agent making repeated observations of an environment undergoing semi-static changes, where features appear and disappear over time, but where these transitions are not necessarily observable. Given observations up to a time $t_N$, our goal is to model and predict the presence or absence of features, also known as feature persistence, at any time \( t \in [t_N, \infty) \). Let \( X_t \in \{0, 1\} \) represent the \emph{persistence} of a feature at time \( t \), where \( X_t = 1 \) indicates presence and \( X_t = 0 \) absence. Given a sequence of noisy observations modeled as Boolean random variables \(\{Y_{t_i}\}_{i=1}^N \subseteq \{0, 1\}^N\), sampled at times \(\{t_i\}_{i=1}^N \in [t_0, \infty)\), the goal is to infer the belief over feature persistence.

\subsection{The Persistence Filter}
\label{sec:persistence}

The \emph{Persistence filter}~\cite{rosen2016towards} is a probabilistic model rooted in Bayesian survival analysis~\cite{ibreahim2005survival} that describes the survival time, or amount of time a feature exists before disappearing, of semi-static features. The persistence model is defined as
\begin{equation}\label{eq:persistence_filter}
    \begin{aligned}
    T &\sim p_T(\cdot), \\
    X_t \mid T &= 
    \begin{cases} 
    1, & t \leq T, \\
    0, & t > T,
    \end{cases} \\
    Y_t \mid X_t &\sim p_{Y_t}(\cdot \mid X_t);
    \end{aligned}
\end{equation}
\noindent
where \(p_T: [0, \infty] \rightarrow [0, \infty]\) is a probability density function denoting the prior over survival time \(T \in [0, \infty)\),  and \( p_{Y_t}(\cdot\mid X_t)\) is the measurement model for some time $t \in [0, \infty)$.  In this model, $X_t = 1$ is equivalent to  $T \geq t$, and we will use them interchangeably. The measurement model is characterized by the probability of missed detections \( P_M = p(Y_t = 0 | X_t = 1) \), and the probability of false alarm \(P_F = p(Y_t = 1 | X_t = 0)\). Further, let the cumulative distribution function (CDF) of \(p_T\) be $F_T(t) \triangleq p(T \leq t) = \int_0^t p_T(\tau)d\tau.$

Given a sequence of noisy observations $\mathcal{Y}_{1:N} \triangleq \{y_{t_i}\}_{i=1}^N$ and the parameters of the measurement model $P_M, P_F \in [0, 1]$, Rosen \etal~\cite{rosen2016towards} derived a closed-form solution to compute the posterior probability $p(X_t = 1 \mid \mathcal{Y}_{1:N})$ at any \emph{present} or \emph{future} time $t \in [t_N, \infty)$
\begin{equation}
    \label{eq:persistence_filter_bayes}
    p(X_t = 1 \mid \mathcal{Y}_{1:N}) = \frac{p(\mathcal{Y}_{1:N} \mid T \geq t)p(T \geq t)}{p(\mathcal{Y}_{1:N})}.
\end{equation}

The general form of the measurement model \( p(\mathcal{Y}_{1:N} \mid T) \), which accounts for both cases \( T \geq t \) and \( T < t \), is given by
\begin{equation}
    \label{eq:persistence_likelihood}
    p(\mathcal{Y}_{1:N} \mid T) = \hspace{-1mm} \prod_{t_i \leq T} \hspace{-1mm} P_M^{1 - y_{t_i}} (1 - P_M)^{y_{t_i}} 
    \hspace{-1mm} \prod_{t_i > T} \hspace{-1mm} P_F^{y_{t_i}} (1 - P_F)^{1 - y_{t_i}}.
\end{equation}
\noindent
Thus, the measurement model \( p(\mathcal{Y}_{1:N} \hspace{-1mm} \mid T \hspace{-1mm} \geq t) \triangleq p(\mathcal{Y}_{1:N} \hspace{-1mm} \mid t_{N}) \) in \eqref{eq:persistence_filter_bayes} for $t_N \hspace{-1mm} > \hspace{-1mm} t_{N-1}, t_{N-2}, \dots$ is recursively updated as
\begin{equation}\label{eq:persistence_likelihood_update}
    \begin{aligned}
    p(\mathcal{Y}_{1:N} \mid T \geq t) = \prod_{i=1}^NP_M^{1 - y_{t_{i}}} (1 - P_M)^{y_{t_{i}}}. 
    \end{aligned}
\end{equation}

As~\eqref{eq:persistence_likelihood} is right-continuous and constant on the interval $[t_i, t_{i+1})$, the evidence $p(\mathcal{Y}_{1:N})$ can be efficiently computed as a sum over disjoint time intervals for all $i \in \{0, 1, \dots, N\}$
\begin{equation}\label{eq:persistence_evidence}
    \begin{aligned}
    p(\mathcal{Y}_{1:N}) &= \sum_{i=0}^N p(\mathcal{Y}_{1:N} \mid t_i)[F_T(t_{i+1}) - F_T(t_i)],
    \end{aligned}
\end{equation}
\noindent
where $t_0 = 0$ and $t_{N+1} = \infty$. Rosen \etal.~\cite{rosen2016towards} showed that~\eqref{eq:persistence_evidence} can be iteratively updated by decomposing it as
\begin{equation}
    \label{eq:persistence_evidence_update}
    p(\mathcal{Y}_{1:N}) = L(\mathcal{Y}_{1:N}) + p(\mathcal{Y}_{1:N} \mid t_N)[1 - F_T(t_N)],
\end{equation}
\noindent
with $L(\mathcal{Y}_{1:N}) \triangleq \sum_{i=0}^{N-1} p(\mathcal{Y}_{1:N} \mid t_i) \left[ F_T(t_{i+1}) - F_T(t_i) \right]$ denoting the \emph{lower partial sum} of the evidence, obtained by excluding the contribution of the \( N^{\text{th}} \) term in~\eqref{eq:persistence_evidence}. In turn, it is easy to iteratively update $L(\mathcal{Y}_{1:N})$ as
\begin{equation}
    \begin{aligned}
        \label{eq:persistence_low_sum_update}
        L(\mathcal{Y}_{1:N}) = & P_F^{y_{t_{N}}} (1 - P_F)^{1 - y_{t_{N}}} \times \\ 
        &(L(\mathcal{Y}_{1:N-1}) + p(\mathcal{Y}_{1:N-1} \mid t_{N-1}) \revised{A_N} ),
    \end{aligned}
\end{equation}
\revised{where $A_N \triangleq F_T(t_{N}) - F_T(t_{N-1})$ is the CDF mass of \(p_T\) in the interval $[t_{N-1}, t_N]$.}

Combining~\eqref{eq:persistence_likelihood_update}, \eqref{eq:persistence_evidence_update}, and \eqref{eq:persistence_low_sum_update}, and noting that the prior follows from $p(T \geq t) = 1 - F_T(t)$, the persistence filter can be iteratively updated in constant time as new observations arrive. Despite its advantages, the persistence filter and its extensions are constrained by three key limitations: (1) there is only a single persistence hypothesis, restricting cases where the feature may exhibit multiple plausible dynamics; (2) once a feature vanishes, it is considered permanently gone, even when it may reappear, and (3) the parameters of the prior are known a priori and not learned from data. In the next section, we present \perpetua, a method that adapts and extends the persistence filter in order to address these limitations.\looseness-1

\section{Perpetua}
The key idea of \perpetua~is to estimate a belief over feature persistence of each feature in the environment using a pair of mixture models: a mixture of \emph{persistence} and \emph{emergence} filters. These models capture the time-dependent state transition probability of appearance $0 \rightarrow 1$ (emergence) and disappearance $1 \rightarrow 0$ (persistence), and are used in combination within a state machine that transitions based on the belief of these filters. At a high level, there are three important subroutines within Perpetua: (1) the mixture models, (2) the Perpetua state machine, and (3) parameter learning. When combined, these subroutines produce a system that can update its belief about the state online and predict arbitrarily far into the future in a principled manner.

\begin{figure}[t]
    \centering
    \includegraphics[width=0.85\linewidth]{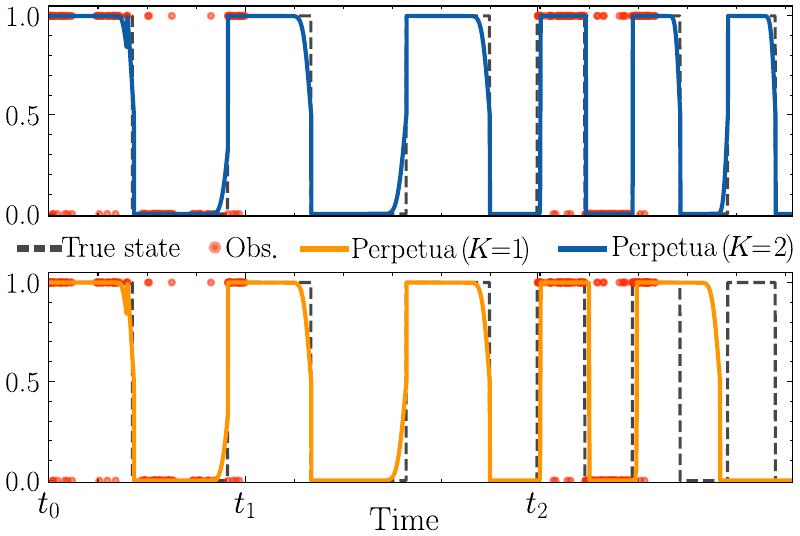}
    \captionof{figure}{
    \footnotesize
    Persistence estimates from Perpetua using mixture sizes of $K\hspace{-0.5mm}=\hspace{-0.5mm}2$ (top) and $K\hspace{-0.5mm}=\hspace{-0.5mm}1$ (bottom). 
    From $t_0$ to $t_1$, persistence estimates adapt to the dynamics given the current noisy observations. 
    Between \( t_1 \) and \( t_2 \), Perpetua uses the component with the largest posterior weight to \emph{predict} feature persistence in the absence of observations. At $t_2$, observations resume. Perpetua ($K\hspace{-0.5mm}=\hspace{-0.5mm}2$) adapts its persistence estimate correctly, switching to the mixture component that best explains the data, while Perpetua ($K\hspace{-0.5mm}=\hspace{-0.5mm}1$) cannot.\looseness-1}
    \vspace{-1.5em}
    \label{fig:perpetua_prediction}
\end{figure}

\subsection{Mixture of Persistence Filters}
\label{sec:death}

Handling multiple persistence hypotheses for the survival time \( T \) is crucial, as the persistence \( X_t \) may have complex dynamics, or a lack of observations may create multiple equally plausible estimates for $T$. This section extends the persistence filter to a mixture model, computing the posterior not only over survival time but also over latent mixture components. This allows selecting the most likely component based on available measurements of a given feature.

We extend the first line in~\eqref{eq:persistence_filter} with an additional latent variable $C  \hspace{-0.5mm} \sim  \hspace{-0.5mm} \text{Categorical}(\pi)$, which models the mixture components, where $C  \in \{1, 2, \dots, K\}$, $\sum_{k=1}^K \pi_k = 1$, and \( p(C = k) = \pi_k \). The mixture of persistence filters is
\begin{equation}\label{eq:mixture_model}
    \begin{aligned}
    T \mid C &= k \sim p_{T_k}(\cdot \mid C = k), 
    \end{aligned}
\end{equation}
\noindent
that, unlike~\eqref{eq:persistence_filter}, has a categorical prior over mixture components $C$, and a conditional prior $ p_{T_k}(\cdot \mid C = k)$ for each mixture component, where $T_k$ is the survival time for component $k$. The rest of the model remains unchanged.\looseness-1

To simplify expressions, we adopt the following notation: $X_t^1 \triangleq (X_t = 1)  $ and $C_k \triangleq (C = k) $. We aim to compute the \emph{joint posterior} \( p(X_t^1, C_k \mid \mathcal{Y}_{1:N}) \) of this mixture model at \( t \in [t_N, \infty) \). Applying Bayes' rule, we have
\begin{equation}\label{eq:mixture_posterior}
    \begin{aligned}
        p(X_t^1, C_k \mid \mathcal{Y}_{1:N}) = \frac{p(X_t^1 \mid C_k, \mathcal{Y}_{1:N}) p(C_k, \mathcal{Y}_{1:N})}{p(\mathcal{Y}_{1:N})},
    \end{aligned}
\end{equation}
\noindent
where $p(C_k, \mathcal{Y}_{1:N})$ is the \emph{joint evidence} and the \emph{conditional posterior} $p(X_t^1 \hspace{-1mm} \mid \hspace{-1mm} C_k, \mathcal{Y}_{1:N})$ is the probability that $X_t = 1$, under component $k$ and given observations $\mathcal{Y}_{1:N}$. Note that each mixture component $C_k$ induces a standard persistence filter through $p(X_t^1 \mid C_k, \mathcal{Y}_{1:N})$. Applying Bayes' rule and the conditional independence $Y_t \perp C \mid X_t$, we decompose the conditional posterior as
\begin{equation}\label{eq:mixture_conditional_posterior}
    \begin{aligned}
        p(X_t^1 \mid C_k, \mathcal{Y}_{1:N}) &= \frac{p(\mathcal{Y}_{1:N} \mid X_t^1)p(X_t^1 \mid C_k)}{p(\mathcal{Y}_{1:N} \mid C_k)}.
    \end{aligned}
\end{equation}

The denominator of~\eqref{eq:mixture_conditional_posterior}, $p(\mathcal{Y}_{1:N} \hspace{-0.5mm} \mid \hspace{-0.5mm} C_k)$, is the \emph{conditional evidence}, which can be computed using a similar approach to~\eqref{eq:persistence_evidence}. Specifically, we recursively update the conditional evidence by first updating the lower partial sum ~\eqref{eq:persistence_low_sum_update} as\looseness-1
\begin{align}
    \label{eq:mixture_low_sum_update}
    L(\mathcal{Y}_{1:N} \mid C_k) = &P_F^{y_{t_N}} (1 - P_F)^{1-y_{t_N}} ( L(\mathcal{Y}_{1:N-1} \mid C_k) \notag \\
    &+  p(\mathcal{Y}_{1:N-1} \mid t_{N-1}) A_{N,k} ),
\end{align}
\noindent 
and plugging in $L(\mathcal{Y}_{1:N} \mid C_k)$ and $F_{T_k}(t_{N}) \hspace{-1mm}\triangleq\hspace{-1mm} \int_0^{t_{N}} p_{T_k}(\tau)d\tau$ into~\eqref{eq:persistence_evidence_update}. \revised{The term $A_{N,k}$ (see \eqref{eq:persistence_low_sum_update}) is computed using the corresponding CDF $F_{T_k}$; both $A_{N,k}$ and $p(\mathcal{Y}_{1:N} \mid C_k)$ are evaluated for all mixture components.} Note that the likelihood $p(\mathcal{Y}_{1:N} \hspace{-1mm} \mid \hspace{-1mm} t_N)$ is recursively updated using~\eqref{eq:persistence_likelihood_update} and the conditional prior can be computed as $p(X_t^1 \mid C_k) = 1 - F_{T_k}(t)$, for all $k \in \{1, 2, \dots, K\}$. Once the conditional evidence \( p(\mathcal{Y}_{1:N} \mid C_k) \) is computed, the joint \( p(C_k, \mathcal{Y}_{1:N}) \) and marginal evidence \( p(\mathcal{Y}_{1:N}) \) can be efficiently obtained as\looseness-1 
\begin{align}
    \label{eq:mixture_joint_evidence}
    &\text{Joint} \quad \rightarrow \quad p(C_k, \mathcal{Y}_{1:N}) = p(\mathcal{Y}_{1:N} \mid C_k) p(C_k), \\
    \label{eq:mixture_marginal_evidence}
    &\text{Marginal} \quad \rightarrow \quad p(\mathcal{Y}_{1:N}) = \sum_{k=1}^K p(\mathcal{Y}_{1:N}, C_k). 
\end{align}
Putting together~\eqref{eq:persistence_likelihood_update}, \eqref{eq:persistence_evidence_update}, and  \eqref{eq:mixture_low_sum_update}-\eqref{eq:mixture_marginal_evidence}, we have a recursive Bayesian estimator for the conditional $p(X_t^1 \mid C_k, \mathcal{Y}_{1:N})$ and joint posterior $p(X_t^1, C_k \mid \mathcal{Y}_{1:N})$ for $t \in [t_N, \infty)$.

By marginalizing the joint posterior $p(X_t^1 \hspace{-1mm} \mid \hspace{-1mm} \mathcal{Y}_{1:N}) = \sum_{k=1}^K p(X_t^1, C_k \mid \mathcal{Y}_{1:N})$, we recover the original persistence filter, where the effects of all mixture components are aggregated. However, due to potential destructive interference between mixture components, we instead use the component with the largest posterior weight. We do this using the posterior weights \( p(C_k \mid \mathcal{Y}_{1:N}) = \frac{p(C_k, \mathcal{Y}_{1:N})}{p(\mathcal{Y}_{1:N})} \), derived from the joint and marginal evidence. Therefore, the probability of the component with the largest posterior weight at time $t \in [t_N, \infty)$ is
\begin{align}
    \label{eq:mixture_mode}
    p(X_t^1 \mid C_{k^*}, \mathcal{Y}_{1:N}) \;\; \text{s.t.} \;\; k^* = \argmax_{k} p(C_k \mid \mathcal{Y}_{1:N}).
\end{align}

The procedure for estimating $X_t$ using a mixture of persistence filters is outlined in Alg.~\ref{alg:mixture_filter}. While mixtures of persistence filters handle multiple persistence hypotheses, once the filter state changes to ``absence,'' it becomes difficult to revert to ``presence'' even with new observations. In the following section, we will address this problem by deriving a related model, called a mixture of emergence filters, that allows us to model the dynamics of feature reappearance.

\begin{algorithm}[t]
\caption{The Mixture of Persistence Filters}
\label{alg:mixture_filter}
\begin{algorithmic}[1]
\Require Observation model $(P_M, P_F)$, CDFs $F_{T_k}(\cdot)$, prior mixture weights $p(C)$, and observations $\{y_{t_i}\}$
\Ensure Probability $p(X_t^1 \mid C_{k^*}, \mathcal{Y}_{1:N})$ for $t \in [t_N, \infty)$. 
\For {$k \in \{1, \dots, K\}$} 
\State $t_0 \gets 0$; $N \gets 0$; $L(\mathcal{Y}_{0} \mid C_k) \gets 0$
\State $p(\mathcal{Y}_{0}) \gets 1$; $p(\mathcal{Y}_{0} \mid t_0) \gets 1$; $p(\mathcal{Y}_{0} \mid C_k) \leftarrow 1$
\EndFor
\Statex \textbf{Update:}
\While{ $\exists$ new observation $y_{t_{N+1}}$}
\For {$k \in \{1, \dots, K\} $}
    \State Compute partial evidence $L(\mathcal{Y}_{1:N+1} \mid C_k)$ via \eqref{eq:mixture_low_sum_update}.
    \State Compute likelihood $p(\mathcal{Y}_{1:N+1} \hspace{-1.0mm} \mid \hspace{-1.0mm} t_{N+1})$ via \eqref{eq:persistence_likelihood_update}.
    \State Compute conditional evidence $p(\mathcal{Y}_{1:N+1} \hspace{-1.2mm} \mid \hspace{-1.2mm} C_k)$ via \eqref{eq:persistence_evidence_update}.
    \State Compute joint evidence $p(C_k, \mathcal{Y}_{1:N+1})$ via \eqref{eq:mixture_joint_evidence}.
\EndFor
\State Compute marginal evidence $p(\mathcal{Y}_{1:N+1})$ via \eqref{eq:mixture_marginal_evidence}.
\State $N \leftarrow (N + 1)$
\Statex \textbf{Predict:} 
\State  For any time \( t \in [t_N, \infty) \), compute the posterior
\Statex \hspace{1.5mm} persistence probability  \( p(X_t^1 \mid C_{k^*}, \mathcal{Y}_{1:N}) \) via \eqref{eq:mixture_mode}.
\EndWhile
\end{algorithmic}
\end{algorithm}

\subsection{Mixture of Emergence Filters}
\label{subsec:birth}

Instead of modeling feature disappearance, the mixture of emergence filters models feature reappearance. The derivation of this model follows similar steps as the mixture of persistence filters with some key differences. We start by updating the conditional $X_t \mid T^{E}$ as
\begin{equation}\label{eq:birth_cond}
    \begin{aligned}
    X_t \mid T^{E} &= 
    \begin{cases} 
    0, & t \leq T^{E} \\
    1, & t > T^{E};
    \end{cases} \\
    \end{aligned}
\end{equation}
\noindent
here, \( T^{E} \) is the emergence time, representing the time it takes for a feature to reappear after disappearing. This formulation is the opposite of the one in~\eqref{eq:persistence_filter}, making the emergence filter the \emph{complement} of the persistence filter.

Using the derivation from \S\ref{sec:death}, we can compute the marginal posterior for the feature's absence (\( X_t = 0 \)) with the mixture of emergence filters: \( p^E(X_T^0 \mid C_k, \mathcal{Y}_{1:N}) \). The posterior for the feature's presence is obtained by taking the complement: \( p^E(X_T^1 \mid C_k, \mathcal{Y}_{1:N}) = 1 - p^E(X_T^0 \mid C_k, \mathcal{Y}_{1:N}) \). To ease notation, we will drop the ``$E$'' superscript. Therefore, to compute $p(X_T^0 \mid C_k, \mathcal{Y}_{1:N})$, the equations for the likelihood \eqref{eq:persistence_likelihood_update}, and lower partial sum \eqref{eq:mixture_low_sum_update} must be updated as:\looseness-1
\begin{align}
    \label{eq:birth_likelihood_update}
    p(\mathcal{Y}_{1:N} \mid t_{N}) &= \prod_{i=1}^N P_F^{y_{t_{i}}} (1 - P_F)^{1 - y_{t_{i}}}  \\
    \label{eq:birth_low_sum_update}
    L(\mathcal{Y}_{1:N} \mid C_k) &= P_M^{1 - y_{t_N}} (1 - P_M)^{y_{t_N}} ( L(\mathcal{Y}_{1:N-1} \mid C_k) \notag \\
    & + p(\mathcal{Y}_{1:N-1} \mid t_{N-1}) B_{N,k} ),
\end{align}
\noindent
\revised{where $B_{N,k}$ is defined like $A_{N,k}$, but uses the CDF, $F_{T_k^E}$, of the emergence prior $p_{T_k^E}(\cdot \mid C_k)$}. The conditional evidence, $p(\mathcal{Y}_{1:N} \mid C_k)$, is recovered by plugging $F_{\tb_k}$,~\eqref{eq:birth_likelihood_update}, and \eqref{eq:birth_low_sum_update} into~\eqref{eq:persistence_evidence_update}. These equations can be incorporated into Alg. \ref{alg:mixture_filter} alongside the cumulative density functions $F_{\tb_k}$ and emergence mixture prior $\pb(C)$ to recursively compute the Bayesian posterior of the emergence model.

The combination of the persistence and emergence mixture models allows us to model both types of state transitions, and operating with mixture models relaxes the assumption of a single dynamical mode. However, we still need to integrate both mixture models to produce a single, coherent estimate for the persistence $X_t$, and do so in a manner that maintains high-quality estimates in the presence of noisy or missing observations. In the next section, we introduce a state machine for performing this integration. 

\subsection{The Perpetua State Machine}
\label{sec:state_machine}

The purpose of the state machine is to control inference on $X_t$ by switching between the persistence and emergence models based on the probability of their heaviest weighted component. When a feature is first observed at time $t_a$, the state machine is initialized in the persistence state. This instantiates a mixture of persistence filters, which starts updating its belief given observations. While in the persistence state, only the persistence mixture is used to estimate $X_t$. If the probability $p(X_t^1 \hspace{-1mm} \mid \hspace{-1mm} C_{k^*}, \mathcal{Y}_{a:b})$, of the persistence model drops below a threshold $\delta_{\text{low}}$ at time $t_b$, Perpetua transitions to the emergence state and instantiates a mixture of emergence filters. The emergence model runs until $p^E(X_t^1 \hspace{-0.8mm} \mid \hspace{-0.8mm} C_{k^*}, \mathcal{Y}_{b:c})$ exceeds a threshold $\delta_{\text{high}}$ at time $t_c$, at which point the state machine returns to the persistence state.\looseness-1

Upon \emph{re-entry} into emergence or persistence states, the respective model is reset, and prior mixture weights updated as\looseness-1
\begin{equation}\label{eq:weights_reset}
    \begin{aligned}
        p^{\text{new}}(C) = \epsilon p(C) + (1 - \epsilon) p(C \mid \mathcal{Y}_{p:q}), \; \; \epsilon \in [0,1],
    \end{aligned}
\end{equation}
\noindent
with $\mathcal{Y}_{p:q}$ denoting the observations taken while the mixture was last active, $p(C)$ the \emph{initial} mixture weights, and $p^{\text{new}}(C)$ the updated weights used to initialize the model after a reset.\looseness-1

The motivation for~\eqref{eq:weights_reset} is two-fold. First, every reset destroys information from previous observations. To alleviate this loss of information, we re-use the posterior weights of the previous model at initialization. Second, incorporating $p(C)$ into~\eqref{eq:weights_reset} prevents mode collapse and maintains the ability to quickly adapt to new modes in the data (see Fig.~\ref{fig:perpetua_prediction}).

Each time a new observation is made, we must assess whether the state of Perpetua has changed between the previous time \( t_{i-1} \) and the current time \( t_i \). To do this, we determine the state change by monitoring the probability of the component with the largest posterior weight of the persistence and emergence mixtures from \( t_{i-1} \) to \( t_i \). During this ``simulation'', Perpetua can switch between the emergence and persistence states multiple times, depending on the temporal distance between \( t_{i-1} \) and \( t_i \), as illustrated in Fig. \ref{fig:perpetua_prediction}.\looseness-1

By combining components from the emergence and persistence models, we can estimate multiple hypotheses by pairing different mixture components. This work uses the component with the largest mixture weight to improve persistence estimates. Fig.~\ref{fig:splash} illustrates how Perpetua estimates the persistence of a feature and all the possible outcomes that can be obtained by combining the mixture components of our models.

\subsection{Parameter Learning}
\label{sec:learning}

One of the strengths of Perpetua is its ability to estimate the parameters of the distributions over $\tb/T$ and $C$ from noisy data. For notational simplicity, we reparameterize $C$ from a categorical variable $C \in \{1, \dots, K\}$ to a one-hot binary vector $C = [C_1, \dots, C_K]^T$, with $p(C_k = 1) = \pi_k$. The element $C_k = 1$ if and only if the result belongs to class $k$, and zero otherwise. We present our derivations for the persistence model but the same steps can be followed to obtain the learning equations of the emergence model. It is worth noting that Dang \etal~\cite{deng2023global} mentioned performing parameter learning over the parameters of \( T \), but we were unable to find details on its derivation.

\textbf{Mixture of Exponential Priors\footnote{The derivation and learning equations for a model with log-normal priors is presented here: \url{https://montrealrobotics.ca/perpetua}.}}. Assume $p_{T_k}$ is an exponential distribution: $p_{T_k}(T \mid C_k = 1) \triangleq \lambda_k \exp{(-\lambda_k T)}$. Since the mixture of persistence filters contains two hidden variables ($C$ and $T$), we use the expectation-maximization algorithm \cite{murphy2022pml1} to estimate the parameters $\Theta \triangleq \{(\lambda_k, \pi_k)\}_{k=1}^K$.

Given an observation sequence $\mathcal{Y}_{1:S}$ with $S >> N$, where a feature may transition multiple times between presence and absence, we assume that these transition times are identifiable. Therefore, we partition our data into $M$ disjoint segments $\mathcal{Y} \triangleq \{\mathcal{Y}_{N_{j}:N_{j+1}}\}_{j=1}^M$, where $N_j$ is the start index of the $j^\text{th}$ segment, $N_{j+1}$ its (non-inclusive) end index, and $N_1 = 1$. Denoting $\mathcal{T} = \{T_j\}_{j=1}^M$ and $\mathcal{C} = \{c_{jk}\}_{j=1, k=1}^{M,K}$, the \emph{complete-data likelihood} can be written as\looseness-1
\begin{equation}\label{eq:learning_likelihood}
\begin{aligned}
    p(\mathcal{Y}, &\mathcal{T}, \mathcal{C}; \Theta) = \prod_{j=1}^M p(\mathcal{Y}_{N_{j}:N_{j+1}} \hspace{-0.5mm} \mid \hspace{-0.5mm} T_j) p(T_j \mid C_j) p(C_j),\\ 
\end{aligned}
\end{equation}
\noindent
with $p(\mathcal{Y}_{N_{j}:N_{j+1}} \hspace{-1mm} \mid \hspace{-1mm} T_j)$ defined by~\eqref{eq:persistence_likelihood}, $p(C_{j}) \triangleq \prod_{k=1}^K \pi_k^{c_{jk}}$, and $p(T_j \mid C_j) = \prod_{k=1}^K [\lambda_k \exp{(-\lambda_k T_j)}]^{c_{jk}}$. Taking the logarithm of~\eqref{eq:learning_likelihood} and then the expectation with respect to \( q^{[u+1]}(\mathcal{T}, \mathcal{C}) \triangleq p(\mathcal{T}, \mathcal{C} \mid \mathcal{Y}; \Theta^{[u]}) \), where \( u \) is the current EM iteration, gives the maximization objective: $\mathcal{L}(q, \Theta) \triangleq \mathbb{E}_{q}[\log p(\mathcal{Y}, \mathcal{T}, \mathcal{C}; \Theta)]$, defined as\looseness-1
%
%
%
\begin{equation}
\begin{aligned}
    \label{eq:learning_elbo}
    \mathcal{L}(q, \Theta) \hspace{-0.5mm} \propto \hspace{-0.5mm} \sum_{j=1}^M \sum_{k=1}^{K} \Big[ \mathbb{E}_{q}[c_{jk}] \left( \log \lambda_k + \log \pi_k\right) \hspace{-0.5mm} - \hspace{-0.5mm} \lambda_k \mathbb{E}_{q}[c_{jk}T_j] \Big ],
\end{aligned}
\end{equation}
\noindent where we dropped the terms that do not depend on $\Theta$.

\textbf{E-Step:} In this step, we fix the parameters $\Theta^{[u]}$ and compute the distribution $q^{[u+1]}$. Although $q$ can be intractable we show it allows for a closed-form solution. From the definition of $q^{[u+1]}(T_j, c_{jk}) \triangleq p(T_j, c_{jk} \mid \mathcal{Y}_{{N_{j}}:N_{j+1}})$, we have
\begin{align}
    \label{eq:learning_q_tc}
    q^{[u+1]}(T_j, c_{jk}) &= \frac{p(\mathcal{Y}_{{N_{j}}:{N_{j+1}}} \mid T_j) p(T_j \mid c_{jk}) p(c_{jk})}{\int_{0}^\infty \sum_{l=1}^K p(\mathcal{Y}_{{N_{j}}:{N_{j+1}}}  \hspace{-1mm} \mid  \hspace{-1mm} \tau) p(\tau  \hspace{-1mm} \mid  \hspace{-1mm} c_{jl}) p(c_{jl}) d\tau}, \\
    \label{eq:learning_q_c}
    q^{[u+1]}(c_{jk}) &= p(c_{jk} \mid \mathcal{Y}_{{N_{j}}:N_{j+1}}).
\end{align}

Note that \eqref{eq:learning_q_c} are the \emph{posterior weights} derived in \S\ref{sec:death}. We define $\phi_{jk}^{[u+1]} \triangleq \mathbb{E}_{q}[c_{jk}] = q^{[u+1]}(c_{jk} = 1)$, and $\psi_{jk}^{[u+1]} \triangleq \mathbb{E}_{q}[c_{jk}T_j]$. Hereafter, we omit the \( u \) superscript to improve readability. By using the same decomposition as in \eqref{eq:persistence_evidence}, and subtracting the last timestamp of set ${j-1}$ from all $t \in \{t_{N_j}, \dots t_{N_{j+1}}\}$ in set $j$, we can compute \( \psi_{jk} \) as
\begin{equation}
    \label{eq:learning_expectation_q_tc}
    \psi_{jk}  \hspace{-0.75mm} = \hspace{-0.5mm} \frac{p(c_{jk} \hspace{-0.5mm} = 1)}{p(\mathcal{Y}_{N_{j}:{N_{j+1}}} \hspace{-0.35mm})} \hspace{-1mm} \sum_{i}^{} p(\mathcal{Y}_{N_{j}:{N_{j+1}}} \hspace{-0.75mm} \mid \hspace{-0.5mm} t_i) \hspace{-0.5mm} \int_{t_i}^{t_{i+1}} \hspace{-3.5mm} p(\tau \hspace{-0.5mm} \mid \hspace{-0.5mm} c_{jk} = 1) \tau  d\tau,
\end{equation}
\noindent with $i \in \{0\} \cup \{N_j, \dots, N_{j+1}\}$. Following \eqref{eq:persistence_evidence}, we set $t_0 = 0$ and $t_{N_{j+1} + 1} = \infty$. For an exponential distribution, it can be shown that $\rho_{ijk} \triangleq \int_{t_i}^{t_{i+1}} p(\tau \mid c_{jk} = 1) \tau  d_\tau $ (with $j$ denoting the observation set) has the closed-form
\begin{equation*}
    \rho_{ijk} = (t_i + \frac{1}{\lambda_k}) \exp({-\lambda_k t_i}) - (t_{i+1} + \frac{1}{\lambda_k}) \exp{(-\lambda_k t_{i+1})}. 
\end{equation*}

\textbf{M-Step:} In this step, we fix the variational distribution $q^{[u+1]}$ and optimize the parameters $\Theta^{[u]}$ by maximizing the objective in~\eqref{eq:learning_elbo}. This leads to the parameters updates
\begin{equation}\label{eq:learning_e_step}
    \lambda^{[u + 1]}_k = \frac{\sum_{j=1}^M \phi_{jk}}{\sum_{j=1}^M \psi_{jk}} \hspace{2mm} \text{ and } \hspace{2mm} \pi^{[u + 1]}_k = \frac{\sum_{j=1}^M \phi_{jk}}{\sum_{j=1}^M\sum_{l=1}^{K} \phi_{jl}}.
\end{equation}
Therefore, to optimize \( \Theta^{[u]} \), we first perform the E-step using~\eqref{eq:learning_q_c} and~\eqref{eq:learning_expectation_q_tc} while keeping the parameters fixed. Then, with \( q^{[u+1]} \) updated, we optimize \( \Theta^{[u+1]} \) using~\eqref{eq:learning_e_step}. This process is repeated for $U$ iterations or until convergence.

\paragraph{Model Selection} To determine the optimal number of components while balancing model complexity, we use the Akaike information criterion (AIC) \cite{akaike1974}:
\begin{equation}  
    \label{eq:aic}  
    AIC = 2p - 2\ln(\hat{L}),  
\end{equation}  

\noindent with \( \hat{L} \triangleq p_{\text{Best}}(\mathcal{Y}_{1:S}) \) denoting the highest evidence achieved by the model, and \( p \) the number of model parameters. The AIC balances goodness of fit and model simplicity by penalizing excessive parameters, helping us to prevent models that overfit the train set.
\section{Results}

We demonstrate that our method, Perpetua, exhibits improved adaptation capabilities compared to baselines while maintaining predictive performance comparable to methods designed to predict persistence. Evaluation is performed in one simulated and one real-world environment with two datasets. We evaluate our method with three metrics: the mean absolute error \(\text{MAE}(f, g) = \frac{1}{(t_1 - t_0)} \int_{t_0}^{t_1} |f(t) - g(t)| \, dt\), where \(f\) is the persistence estimate and \(g\) the ground truth; balanced accuracy \(\text{B-Acc} = \frac{1}{2} \left( \frac{\text{TP}}{\text{TP} + \text{FN}} + \frac{\text{TN}}{\text{TN} + \text{FP}} \right)\), which accounts for class imbalance by equally weighting true positives and true negatives; and F1 score \(\text{F1} = \frac{2\text{TP}}{2\text{TP} + \text{FP} + \text{FN}}\), to evaluate prediction reliability by balancing precision and recall. Here, TP, FP, TN and FN denote true/false positives, and true/false negatives, respectively. We threshold persistence estimates at 0.5 to compute B-Acc and F1 score.

Four baselines are used throughout the evaluation:
\begin{itemize}
    \item \textbf{FreMen} \cite{krajnik2017fremen}. Following the authors, we fit the model using 1000 Fourier coefficients but determine the optimal number for prediction via a held-out validation set. The model is periodically re-fit in all experiments.
    \item \textbf{ARMA} \cite{wang2020arma, wang2024arma}. Following the authors' procedure, we determine the model order by minimizing the AIC. As no implementation was available, we have re-implemented it ourselves. The model is periodically re-fit.
    \item \textbf{PF} \cite{rosen2016towards}. We extend the persistence filter with our mixture formulation, using a mixture of exponential distributions. The model parameters are learned from data following the procedure outlined in \S\ref{sec:learning}.
    \item \textbf{PF LSTE} \cite{deng2023global}. The persistence filter with long short-term exponential memory. For consistency, we also extend this model as we did for the PF. Since no implementation was available, we implemented it ourselves.
\end{itemize}

We set \(\delta_\text{low} \hspace{-0.5mm}=\hspace{-0.5mm} 0.05\), \(\delta_\text{high} \hspace{-0.5mm}=\hspace{-0.5mm} 0.95\) \revised{to avoid ambiguous switching, and empirically found} \(\epsilon \hspace{-0.5mm}=\hspace{-0.5mm} 0.1\) \revised{to perform well; these values were fixed across all features and experiments.} Perpetua is evaluated using two mixture distributions: exponential (Exp) and log-normal (Log-N). \revised{We chose these distributions as both admit closed-form parameter solutions; however, Perpetua can be used with any distribution supported on $\mathbb{R}^{\geq 0}$.} The \texttt{ruptures} library~\cite{truong2020ruptures} is used during training to detect change points in the data (see \S\ref{sec:learning}). The number of mixture components is determined using the AIC (\ref{eq:aic}) with $k \in \{1,\dots,5\}$. Our models are trained for up to 250 iterations with uninformed initialization; $P_M$ and $P_F$ for our method and filter-based baselines are determined from dataset statistics.\looseness-1

\subsection{Simulation in Room Environment}
\label{sec:sim}

\begin{figure}[t]
 \centering\includegraphics[width=0.49\linewidth]{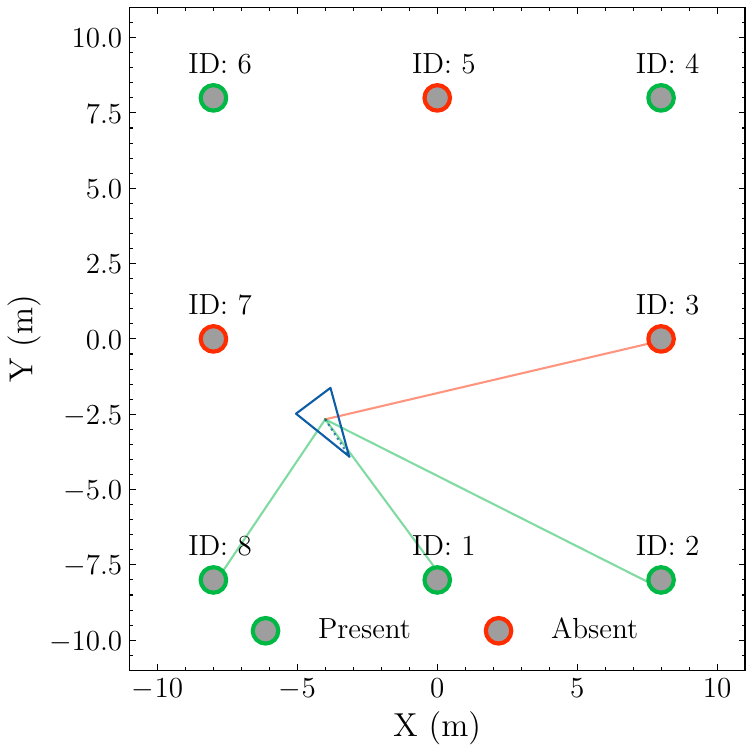}
   \includegraphics[width=0.49\linewidth]{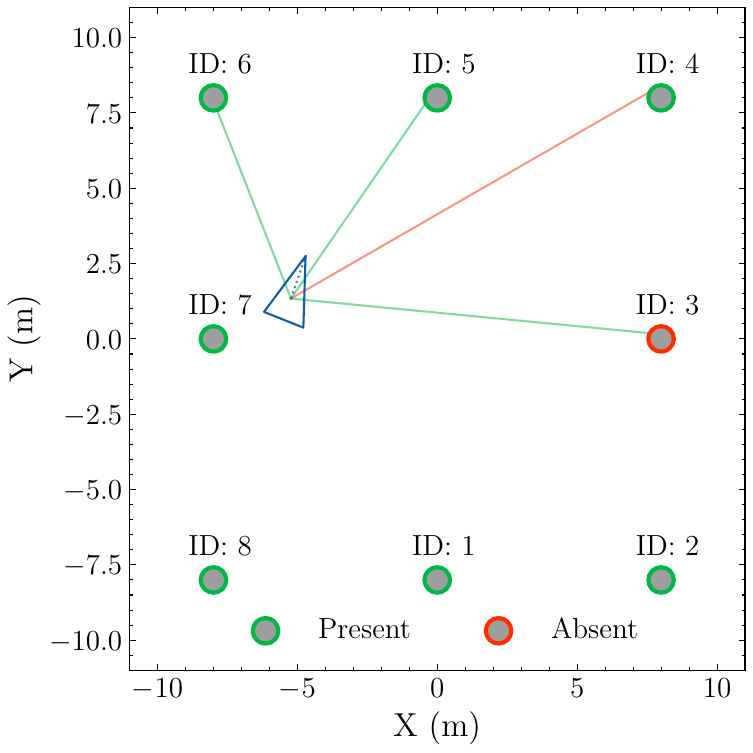}
        \vspace{-1.5em}
    \captionof{figure}{\footnotesize Robot collecting data in the simulated room. (left) train; (right) test.}
    \vspace{-1.5em}
    \label{fig:sim_room}
\end{figure}

The first evaluation, conducted in simulation, compares the ability of Perpetua and baselines to estimate persistence. Here, a robot navigates a room with eight landmarks \revised{(features)}, where four are static and four are semi-static (see Fig.~\ref{fig:sim_room}). Semi-static landmarks have different appearance and disappearance times, ranging from 1min to 20min, sampled from a mixture of log-normal distributions. Landmarks can have up to three distinct appearance and disappearance times.

During data collection, the robot navigates counter-clockwise for 12 hours, while during testing the robot moves clockwise for 3 hours. To avoid measuring at regular intervals, the robot's speed is varied between \(0.5 \frac{m}{s}\) and \(2.5 \frac{m}{s}\) every five minutes. Both training and test sets have observation noise $P_M \hspace{-0.5mm} = \hspace{-0.5mm} P_F \hspace{-0.5mm} = 0.1$. Ground-truth is computed every 0.5 seconds, including intervals where features are not observed.\looseness-1

Since all methods based on the persistence filter can leverage noisy test-time observations, we use the following evaluation procedure: Given a time $t$ in the test set, we allow the model to process all data up to $t$ and then query the model to estimate persistence at time $t + \Delta t$. We call $\Delta t$ the prediction time. For methods requiring re-fitting, such as FreMen and ARMA, we re-fit the model after consuming data up to $t$, following the same procedure. While in some cases re-fitting may not be possible in real-time, we chose this approach to ensure a fair comparison.

\begin{figure*}[ht!]
    \centering
    \includegraphics[width=\textwidth]{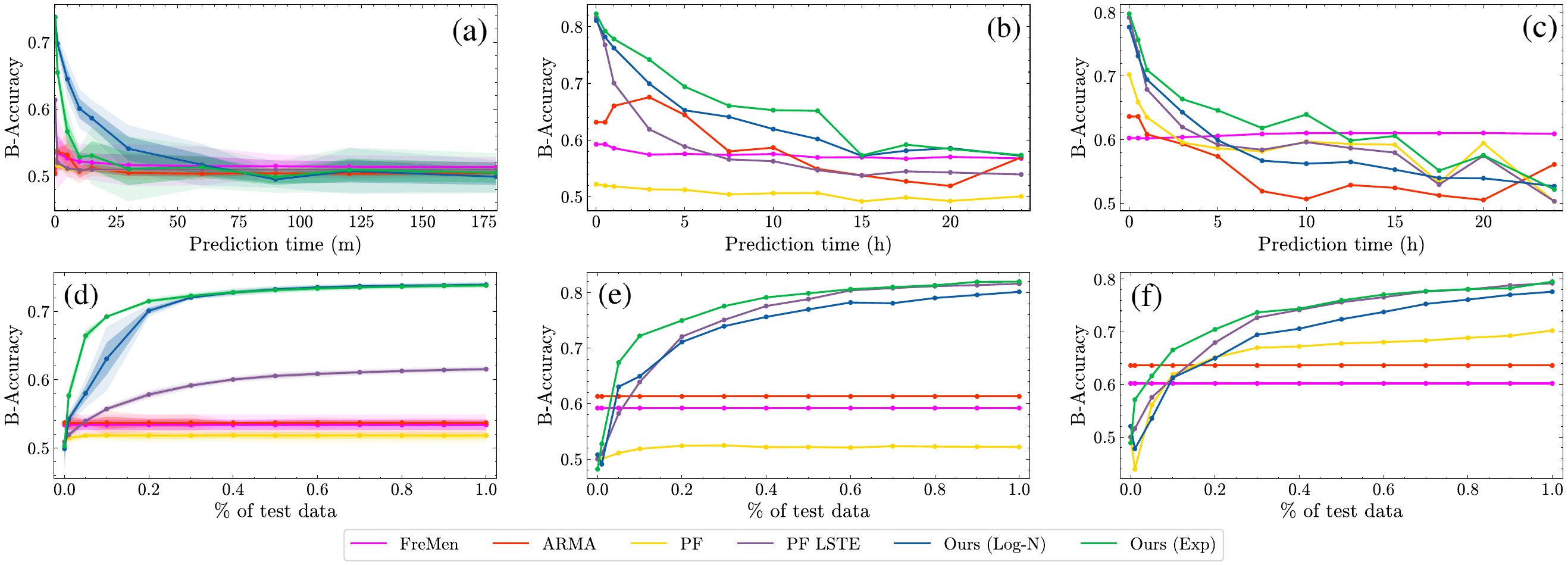}
    \vspace{-0.75em}
    \captionof{figure}{\footnotesize Prediction (a-c) and data sparsity ablation (d-f) in the simulated room (a,d) and parking lot datasets, UFPR04: (b,e), UFPR05: (c,f). The top row shows the balanced accuracy of all methods when predicting feature persistence for a prediction time ($\Delta t$) in the future. The bottom row presents an ablation study on data sparsity, evaluating filter-based methods when only a random subset of the noisy test data is available. These results highlight the robustness of filtering methods, and Perpetua in particular, to limited data. The best predictive versions of FreMen and ARMA are included for reference.}
    \vspace{-1.5em}
    \label{fig:benchmarks}
\end{figure*}

Table~\ref{tab:sim_room} (see also Fig.~\ref{fig:benchmarks}(a)) presents the results for all methods in the simulated room experiment. The results correspond to prediction times $\Delta t \in [0, 3\text{hr}]$ where each model achieved the highest B-Acc, \revised{i.e., the maximum value shown in Fig.~\ref{fig:benchmarks}(a). This choice was made to highlight the best attainable performance of each method.} Note that maximizing B-Acc may not simultaneously achieve the lowest MAE or highest F1. In general, Perpetua outperforms the baselines across all metrics. The two bottom rows of Table~\ref{tab:sim_room} show Perpetua's performance when given the ground truth parameters governing the dynamics, demonstrating that the parameters Perpetua learns in the uninformed case are highly accurate.\looseness-1

\begin{table}[htpb!]
\centering
\renewcommand{\arraystretch}{1.25}
\setlength{\tabcolsep}{3.15pt}
\captionsetup{width=\columnwidth}
\vspace{-0.6em}
\captionof{table}{\footnotesize Results in Room Environment. All metrics are averaged over five random seeds. Shaded rows show ground-truth-informed Perpetua.} 
\vspace{-0.5em}
\begin{tabular}{lccc}
\toprule
Method & MAE $\downarrow$ & B-Acc $\uparrow$ & F1 $\uparrow$ \\
\midrule
FreMen \cite{krajnik2017fremen} & $0.244 \pm 0.015$ & $0.534 \pm 0.027$ & $0.700 \pm 0.055$  \\
ARMA \cite{wang2020arma}  & $0.232 \pm 0.003$ & $0.537 \pm 0.006$ & $0.691 \pm 0.012$  \\
\multirow{1}{*}{PF \cite{rosen2016towards}} & $0.125 \pm 0.007$ & $0.519 \pm 0.005$ & $0.569 \pm 0.016$  \\ 
\multirow{1}{*}{PF LSTE \cite{deng2023global}} & $0.110 \pm 0.002$ & $0.614 \pm 0.004$ & $0.806 \pm 0.007$  \\ 
\multirow{1}{*}{Ours (Exp)} & $0.015 \pm 0.001$ & $0.737 \pm 0.002$ & $0.981 \pm 0.003$  \\
\multirow{1}{*}{Ours (Log-N)} & $\mb{0.010 \pm 0.001}$ & $\mb{0.738 \pm 0.003}$ & $\mb{0.983 \pm 0.004}$  \\
\midrule
\rowcolor{gray!20}
Ours GT (Exp)     & $0.015 \pm 0.001$ & $0.737 \pm 0.002$ & $0.981 \pm 0.003$  \\
\rowcolor{gray!20}
Ours GT (Log-N)     & $0.011 \pm 0.001$ & $0.736 \pm 0.001$ & $0.980 \pm 0.002$  \\
\bottomrule
\end{tabular}
\label{tab:sim_room}
\vspace{-0.75em}
\end{table}

We also evaluate the robustness of the filter-based methods to data sparsity. Here, persistence estimators receive only a random subset of noisy observations from the test set, and prediction time $\Delta t = 0$. As shown in Fig.~\ref{fig:benchmarks}~(d), Perpetua outperforms the baselines using 5\% of the test observations, and starts to plateau at 20\%. We present the best FreMen and ARMA models from the previous experiment for reference.

\subsection{Parking Lot Dataset}

This experiment tests persistence estimators when learning model parameters from real-world data where priors are unavailable. For this, we use the parking lot dataset from Almeida \etal~\cite{almeida2015pklot}, which consists of 12,427 images of parking lots captured under varying environmental conditions. These images were taken in the parking lots of the Federal University of Parana (UFPR), with observations recorded every five minutes for over 30 days. This evaluation focuses on the UFPR04 and UFPR05 subsets, which represent different views of the same parking lot captured from the fourth and fifth floors of the UFPR building. The UFPR04 set contains 28 parking spaces (tracked features), while UFPR05 has 45.

\begin{figure}[t]
 \vspace{0.65em}
 \centering\includegraphics[width=0.49\linewidth]{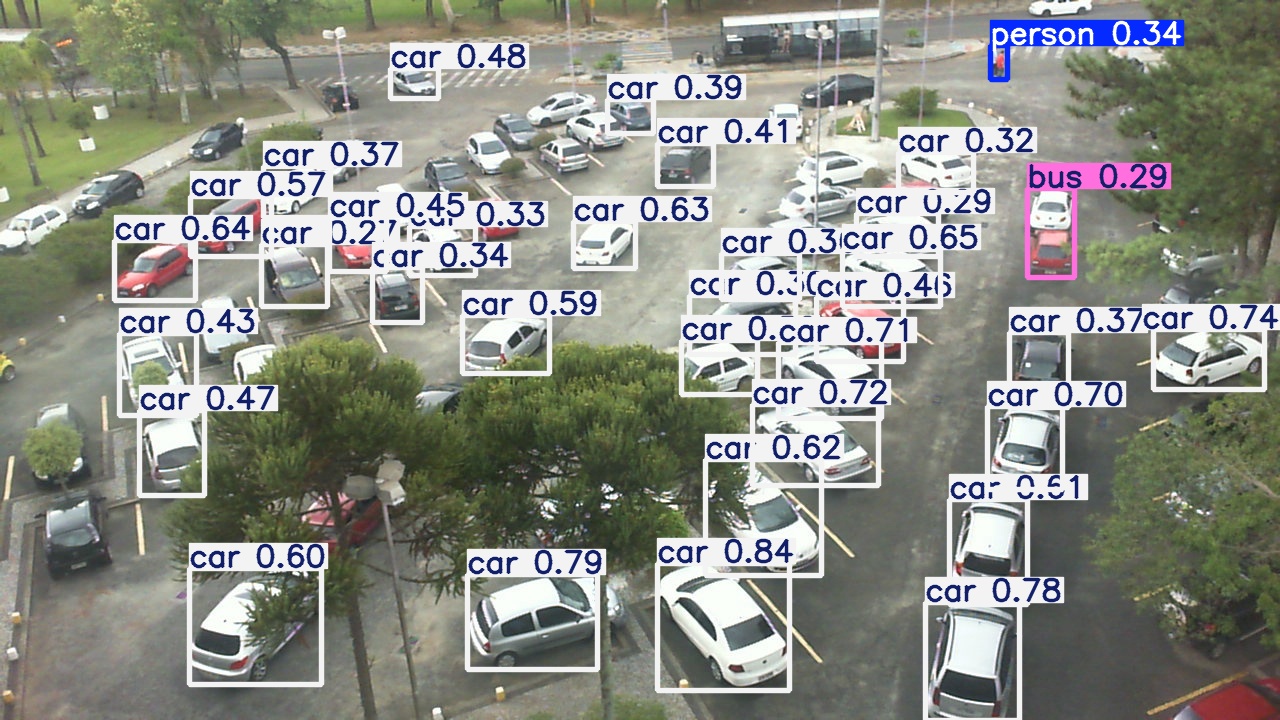}
   \includegraphics[width=0.49\linewidth]{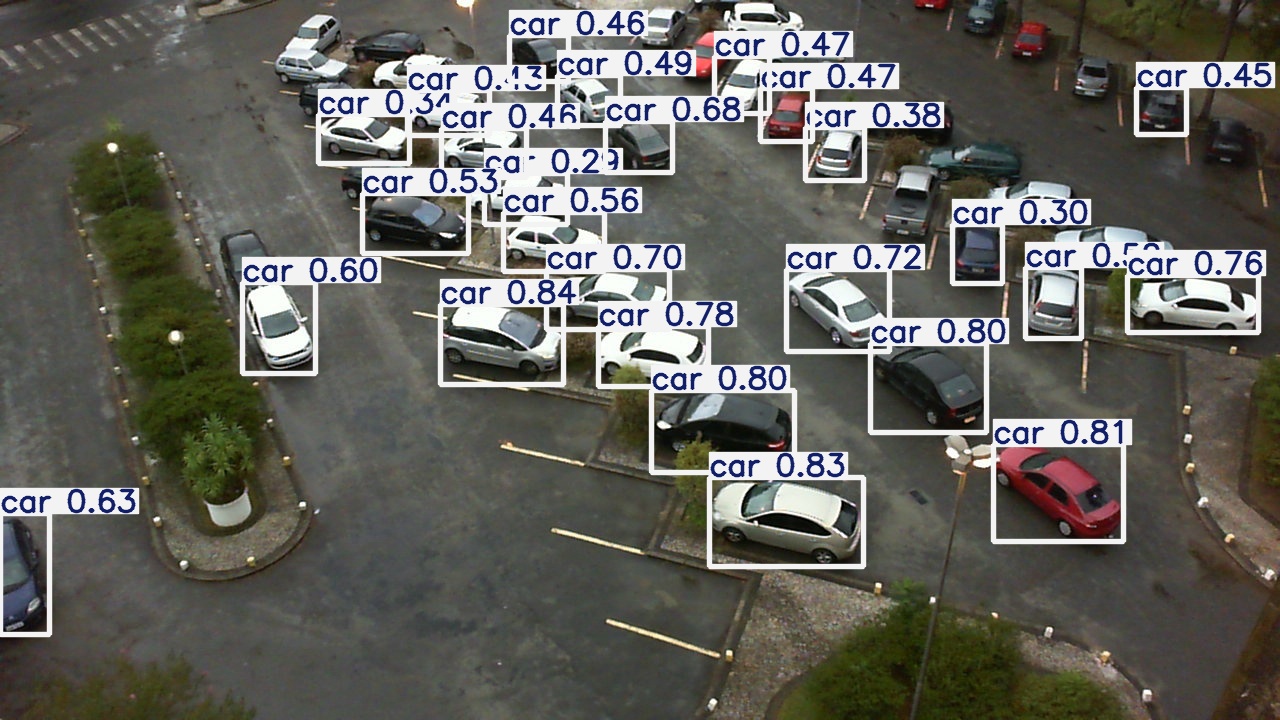}
        \vspace{-1.5em}
    \captionof{figure}{\footnotesize YOLOv11 applied to UFPR04 (left) and UFPR05 (right).}
    \vspace{-1.35em}
    \label{fig:pklot}
\end{figure}

To match real-world conditions, we process the dataset using an off-the-shelf object detector, YOLOv11~\cite{yolo11_ultralytics}, to obtain observations. Example detections are shown in Fig.~\ref{fig:pklot}, where some vehicles are not detected, introducing noise into the observations. YOLOv11 achieves an accuracy of 85.5\% on UFPR04 and 79.9\% on UFPR05. We follow the same evaluation procedure described in \S\ref{sec:sim} and use the last 30\% of data as the test set.

The results, shown in Fig.~\ref{fig:benchmarks}~(b-c), demonstrate that Perpetua outperforms all methods in the UFPR04 set across all prediction times. In the UFPR05 set, Perpetua (Exp) outperforms the baselines for prediction time $\Delta t \leq 10\text{hr}$, beyond which fully predictive methods such as FreMen and ARMA surpass all filter-based methods, with Perpetua remaining the most accurate filter-based approach. These results suggest that fitting a mixture of exponential distributions may be easier than a mixture of log-normals. Table~\ref{tab:pklot} presents the most accurate models for each method over all prediction times.\looseness-1

\begin{table}[ht!]
\centering
\renewcommand{\arraystretch}{1.25}
\setlength{\tabcolsep}{3.5pt}
\captionsetup{width=\columnwidth}
\captionof{table}{\footnotesize Results in Parking Lot dataset. UFPR04 has 28 parking spaces (tracked features) and UFPR05 contains 45.}
\vspace{-0.5em}
\begin{tabular}{lcccccc}
\toprule
\multirow{2}[3]{*}{Method}  &  \multicolumn{3}{c}{UFPR04} &  \multicolumn{3}{c}{UFPR05} \\ 
\cmidrule(lr){2-4} \cmidrule(lr){5-7}
& MAE $\downarrow$ & B-Acc $\uparrow$ & F1 $\uparrow$ & MAE $\downarrow$ & B-Acc $\uparrow$ & F1 $\uparrow$ \\
\midrule
FreMen \cite{krajnik2017fremen}  & 0.389 & 0.592 & 0.470 & 0.388 & 0.602 & 0.560 \\
\multirow{1}{*}{ARMA \cite{wang2020arma}}
                 & 0.409 & 0.613 & 0.451 & 0.396 & 0.636 & 0.536 \\ 
\multirow{1}{*}{PF \cite{rosen2016towards}}
                 & 0.539 & 0.512 & 0.067 & 0.421 & 0.702 & 0.528 \\ 
\multirow{1}{*}{PF LSTE \cite{deng2023global}}
                 & 0.208 & 0.813 & 0.733 & \textbf{0.285} & 0.792 & 0.738 \\ 
\multirow{1}{*}{Ours (Exp)}
                 & \textbf{0.201} & \textbf{0.823} & \textbf{0.745} & 0.359 & \textbf{0.798} & 0.785 \\
\multirow{1}{*}{Ours (Log-N)}
                 & 0.228 & 0.811 & 0.735 & 0.359 & 0.777 & \textbf{0.794}\\
\bottomrule
\end{tabular}
\label{tab:pklot}
\vspace{-1.5em}
\end{table}

\begin{figure}[htpb!]
    \centering
    \includegraphics[width=0.95\linewidth]{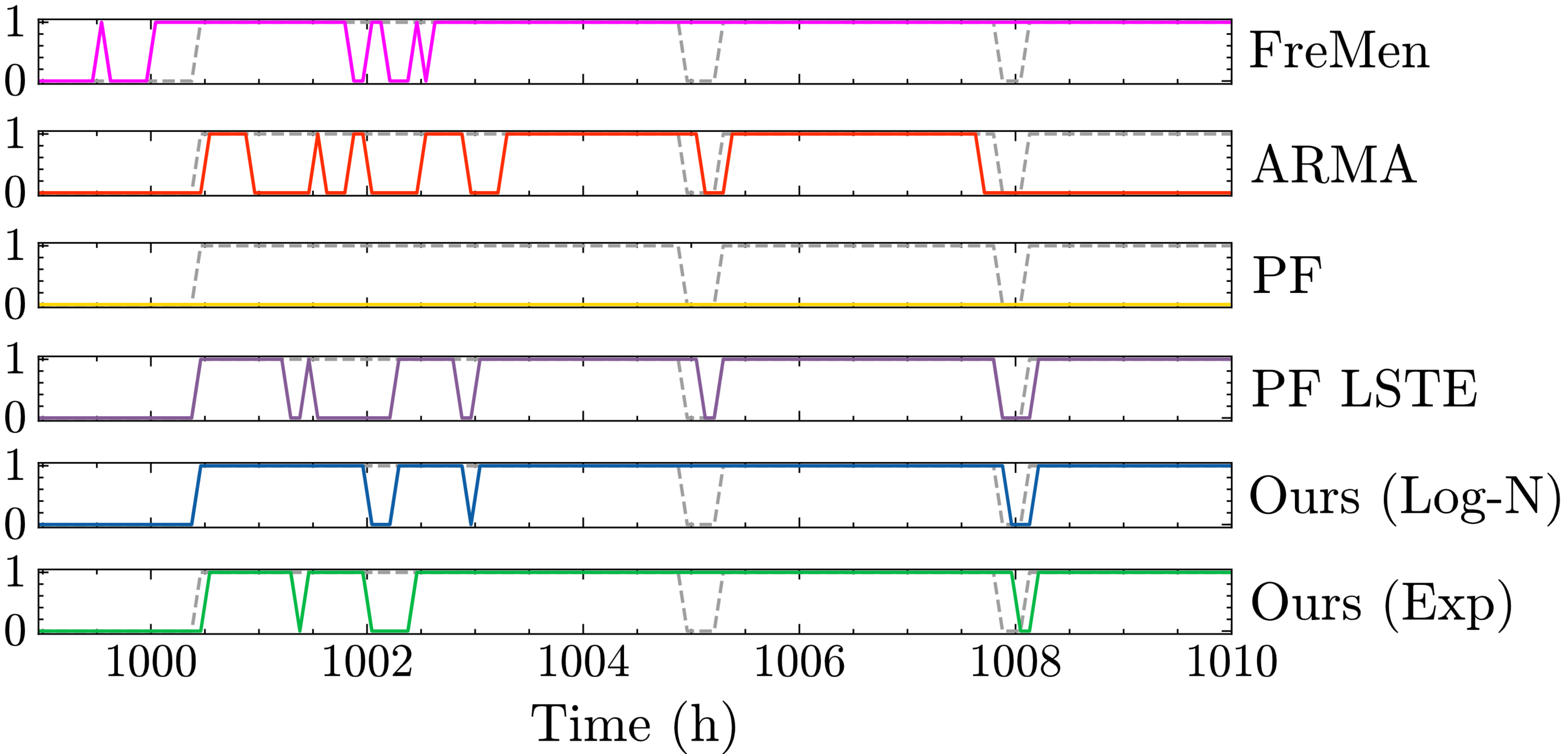}
    \vspace{-0.5em}
    \captionof{figure}{\footnotesize Example results obtained by our method and baselines when estimating the persistence of a feature in the UFPR04 data set. Here, we threshold the persistence estimates of all methods at 0.5.}
    \vspace{-0.5em}
    \label{fig:all_filters}
\end{figure}

In Fig.~\ref{fig:benchmarks}~(e-f), we present the results of the data sparsity ablation. Perpetua (Exp) outperforms the baselines using only 10\% of noisy test-time data. Although PF LSTE is similarly robust to data sparsity, its overall prediction performance (MAE, B-Acc, F1) is weaker than Perpetua. Persistence estimates for UFPR04 are illustrated in Fig.~\ref{fig:all_filters}. 

\revised{While we do not explicitly address spurious observations from data association errors, Perpetua’s probabilistic formulation can account for them by adjusting $P_M$ and $P_F$ based on front-end reliability. In real-world settings, model parameters may initially be unidentifiable due to sparse observations, particularly for newly observed features, but they can be refined over time as the agent gathers more data.}\looseness-1

\textit{Implementation and Hardware Details.} We implement Perpetua using JAX \cite{jax2018github}, on a Ryzen 1950X CPU with 32GB of RAM. Training a mixture model with exponential priors on 12,500 data points takes 0.058 seconds per iteration, and 0.170 seconds for the log-normal prior.

\section{Conclusion}

This paper presents Perpetua, a method for estimating feature persistence that models and predicts semi-static dynamics with robust online adaptation capabilities. We show Perpetua's superior adaptation and predictive capabilities on both real and simulated data, and present additional results on robustness to missing observations. In the future, we plan to explore updating the persistence priors of our mixture models dynamically using large vision-language models. Finally, we intend to further evaluate Perpetua in complex sequential reasoning domains within long-horizon planning tasks. 

\bibliographystyle{IEEEtran}
\bibliography{IEEEabrv.bib,  short-string.bib, bibliography.bib}

\begin{thebibliography}{10}
\providecommand{\url}[1]{#1}
\csname url@samestyle\endcsname
\providecommand{\newblock}{\relax}
\providecommand{\bibinfo}[2]{#2}
\providecommand{\BIBentrySTDinterwordspacing}{\spaceskip=0pt\relax}
\providecommand{\BIBentryALTinterwordstretchfactor}{4}
\providecommand{\BIBentryALTinterwordspacing}{\spaceskip=\fontdimen2\font plus
\BIBentryALTinterwordstretchfactor\fontdimen3\font minus \fontdimen4\font\relax}
\providecommand{\BIBforeignlanguage}[2]{{%
\expandafter\ifx\csname l@#1\endcsname\relax
\typeout{** WARNING: IEEEtran.bst: No hyphenation pattern has been}%
\typeout{** loaded for the language `#1'. Using the pattern for}%
\typeout{** the default language instead.}%
\else
\language=\csname l@#1\endcsname
\fi
#2}}
\providecommand{\BIBdecl}{\relax}
\BIBdecl

\bibitem{adkins2022probabilistic}
A.~Adkins, T.~Chen, and J.~Biswas, ``Probabilistic object maps for long-term robot localization,'' in \emph{Proc. {IEEE}/{RSJ} Int. Conf. Intell. Robots and Syst.}, 2022.

\bibitem{hashemifar2020visual}
Z.~Hashemifar and K.~Dantu, ``Practical persistence reasoning in visual slam,'' in \emph{Proc. {IEEE} Int. Conf. Robot. and Automation}, 2020.

\bibitem{schmid2024khronos}
L.~Schmid, M.~Abate, Y.~Chang, and L.~Carlone, ``Khronos: A unified approach for spatio-temporal metric-semantic slam in dynamic environments,'' in \emph{Proc. of Robotics: Science and Systems (RSS)}, 2024.

\bibitem{nashed2016curating}
S.~Nashed and J.~Biswas, ``Curating long-term vector maps,'' in \emph{Proc. {IEEE}/{RSJ} Int. Conf. Intell. Robots and Syst.}, 2016.

\bibitem{krajnik2017fremen}
T.~Krajník, J.~P. Fentanes, J.~M. Santos, and T.~Duckett, ``Fremen: Frequency map enhancement for long-term mobile robot autonomy in changing environments,'' \emph{{IEEE} Trans. on Robot.}, 2017.

\bibitem{qian2024closing}
J.~Qian, S.~Zhou, N.~J. Ren, V.~Chatrath, and A.~P. Schoellig, ``Closing the perception-action loop for semantically safe navigation in semi-static environments,'' in \emph{Proc. {IEEE} Int. Conf. Robot. and Automation}, 2024.

\bibitem{nardi2020nav}
L.~Nardi and C.~Stachniss, ``Long-term robot navigation in indoor environments estimating patterns in traversability changes,'' in \emph{Proc. {IEEE} Int. Conf. Robot. and Automation}, 2020.

\bibitem{krajnik2021khrono}
T.~Krajn\'{\i}k \emph{et~al.}, ``Chronorobotics: Representing the structure of time for service robots,'' in \emph{Proceedings of the International Symposium on Computer Science and Intelligent Control}, 2021.

\bibitem{wang2020arma}
L.~Wang, W.~Chen, and J.~Wang, ``Long-term localization with time series map prediction for mobile robots in dynamic environments,'' in \emph{Proc. {IEEE}/{RSJ} Int. Conf. Intell. Robots and Syst.}, 2020.

\bibitem{rosen2016towards}
D.~Rosen, J.~Mason, and J.~Leonard, ``Towards lifelong feature-based mapping in semi-static environments,'' in \emph{Proc. {IEEE} Int. Conf. Robot. and Automation}, 2016.

\bibitem{nobre2018online}
F.~Nobre, C.~Heckman, P.~Ozog, R.~W. Wolcott, and J.~M. Walls, ``Online probabilistic change detection in feature-based maps,'' in \emph{Proc. {IEEE} Int. Conf. Robot. and Automation}, 2018.

\bibitem{biber2005dynamicMaps}
P.~Biber and T.~Duckett, ``Dynamic maps for long-term operation of mobile service robots,'' in \emph{Robotics: Science and Systems}, 2005.

\bibitem{cadena2016past}
C.~Cadena \emph{et~al.}, ``Past, present, and future of simultaneous localization and mapping: Toward the robust-perception age,'' \emph{{IEEE} Trans. on Robot.}, 2016.

\bibitem{schmid2022panoptic}
L.~Schmid \emph{et~al.}, ``Panoptic multi-{TSDFs}: {A} flexible representation for online multi-resolution volumetric mapping and long-term dynamic scene consistency,'' in \emph{Proc. {IEEE} Int. Conf. Robot. and Automation}, 2022.

\bibitem{saarinen2012markov}
J.~Saarinen, H.~Andreasson, and A.~J. Lilienthal, ``Independent markov chain occupancy grid maps for representation of dynamic environment,'' in \emph{Proc. {IEEE}/{RSJ} Int. Conf. Intell. Robots and Syst.}, 2012.

\bibitem{tipaldi2013lifelong}
G.~D. Tipaldi, D.~Meyer-Delius, and W.~Burgard, ``Lifelong localization in changing environments,'' \emph{The International Journal of Robotics Research}, 2013.

\bibitem{fu2023neuse}
J.~Fu, Y.~Du, K.~Singh, J.~B. Tenenbaum, and J.~J. Leonard, ``Neu{SE}: Neural {SE}(3)-equivariant embedding for consistent spatial understanding with objects,'' in \emph{Robotics: Science and Systems (RSS)}, 2023.

\bibitem{looper20233d}
S.~Looper, J.~Rodriguez-Puigvert, R.~Siegwart, C.~Cadena, and L.~Schmid, ``{3D VSG}: Long-term semantic scene change prediction through 3d variable scene graphs,'' in \emph{Proc. {IEEE} Int. Conf. Robot. and Automation}, 2023.

\bibitem{thomas2023foreseeable}
H.~Thomas, J.~Zhang, and T.~D. Barfoot, ``The foreseeable future: Self-supervised learning to predict dynamic scenes for indoor navigation,'' \emph{{IEEE} Trans. on Robot.}, 2023.

\bibitem{krajnik2019warped}
T.~Krajn{\'\i}k \emph{et~al.}, ``Warped hypertime representations for long-term autonomy of mobile robots,'' \emph{{IEEE} Robot. and Automation Lett.}, 2019.

\bibitem{guizilini2019hilbert}
V.~Guizilini, R.~Senanayake, and F.~Ramos, ``Dynamic hilbert maps: Real-time occupancy predictions in changing environments,'' in \emph{Proc. {IEEE} Int. Conf. Robot. and Automation}, 2019.

\bibitem{wang2024arma}
Y.~Wang, Y.~Fan, J.~Wang, and W.~Chen, ``Long-term navigation for autonomous robots based on spatio-temporal map prediction,'' \emph{Robotics and Autonomous Systems}, 2024.

\bibitem{deng2023global}
T.~Deng, H.~Xie, J.~Wang, and W.~Chen, ``Long-term visual simultaneous localization and mapping: Using a bayesian persistence filter-based global map prediction,'' \emph{IEEE Robotics \& Automation Magazine}, 2023.

\bibitem{ibreahim2005survival}
J.~Ibrahim, M.~H. Chen, and D.~Sinha, \emph{Bayesian Survival Analysis}.\hskip 1em plus 0.5em minus 0.4em\relax Springer, 2005.

\bibitem{murphy2022pml1}
K.~P. Murphy, \emph{Probabilistic Machine Learning: An introduction}.\hskip 1em plus 0.5em minus 0.4em\relax MIT Press, 2022.

\bibitem{akaike1974}
H.~Akaike, ``A new look at the statistical model identification,'' in \emph{Selected Papers of Hirotugu Akaike}.\hskip 1em plus 0.5em minus 0.4em\relax Springer, 1974.

\bibitem{truong2020ruptures}
C.~Truong, L.~Oudre, and N.~Vayatis, ``Selective review of offline change point detection methods,'' \emph{Signal Processing}, 2020.

\bibitem{almeida2015pklot}
P.~Almeida, L.~S. Oliveira, E.~Silva~Jr, A.~Britto~Jr, and A.~Koerich, ``Pklot -- a robust dataset for parking lot classification,'' \emph{Expert Systems with Applications}, 2015.

\bibitem{yolo11_ultralytics}
\BIBentryALTinterwordspacing
G.~Jocher and J.~Qiu, ``Ultralytics yolo11,'' 2024. [Online]. Available: \url{https://github.com/ultralytics/ultralytics}
\BIBentrySTDinterwordspacing

\bibitem{jax2018github}
\BIBentryALTinterwordspacing
J.~Bradbury \emph{et~al.}, ``{JAX}: composable transformations of {P}ython+{N}um{P}y programs,'' 2018. [Online]. Available: \url{http://github.com/jax-ml/jax}
\BIBentrySTDinterwordspacing

\end{thebibliography}

\end{document}